\documentclass[10pt,twocolumn,letterpaper]{article}

\usepackage{cvpr}
\usepackage{times}
\usepackage{epsfig}
\usepackage{graphicx}
\usepackage{amsmath}
\usepackage{amssymb}

\graphicspath{ {images/} }
\usepackage[breaklinks=true,bookmarks=false]{hyperref}

\cvprfinalcopy 

\def\cvprPaperID{****} 

\begin{document}

\title{Self-Driving Car Steering Angle Prediction Based on Image Recognition}

\author{Shuyang Du\\
{\tt\small shuyangd@stanford.edu}
\and
Haoli Guo\\
{\tt\small haoliguo@stanford.edu}
\and
Andrew Simpson\\
{\tt\small asimpso8@stanford.edu}
}

\maketitle

\begin{abstract}
Self-driving vehicles have expanded dramatically over the last few years. Udacity has release a dataset containing, among other data, a set of images with the steering angle captured during driving. The Udacity challenge aimed to predict steering angle based on only the provided images. 

We explore two different models to perform high quality prediction of steering angles based on images using different deep learning techniques including Transfer Learning, 3D CNN, LSTM and ResNet. If the Udacity challenge was still ongoing, both of our models would have placed in the top ten of all entries. 

\end{abstract}

\section{Introduction}
Self-driving vehicles are going to be of enormous economic impact over the coming decade. Creating models that meet or exceed the ability of a human driver could save thousands of lives a year. Udacity has an ongoing challenge to create an open source self-driving car \cite{udacity}. In their second challenge Udacity released a dataset of images taken while driving along with the corresponding steering angle and ancillary sensor data for a training set (left, right, and center cameras with interpolated angles based on camera angle). The goal of the challenge was to find a model that, given an image taken while driving, will minimize the RMSE (root mean square error) between what the model predicts and the actual steering angle produced by a human driver. In this project, we explore a variety of techniques including 3D convolutional neural networks, recurrent neural networks using LSTM, ResNets, etc. to output a predicted steering angle in numerical values.

The motivation of the project is to eliminate the need for hand-coding rules and instead create a system that learns how to drive by observing. Predicting steering angle is one important part of the end-to-end approach to self-driving car and would allow us to explore the full power of neural networks. For example, using only steering angle as the training signal, deep neural networks can automatically extract features to help position the road to make the prediction.

The two models that will be discussed are a model that used 3D convolutional layers followed by recurrent layers using LSTM (long short term memory). This model will explore how temporal information is used to predict steering angle. Both 3D convolutional layers and recurrent layers make use of temporal information. The second model to be discussed uses transfer learning (using lower layers of a pre-trained model) by using a high quality model trained on another dataset. Transfer learning helps to mitigate the amount of training data needed to create a high quality model.

\section{Related Work}

Using a neural network for autonomous vehicle navigation was pioneered by Pomerleau (1989) \cite{pomerleau1989alvinn} who built the Autonomous Land Vehicle in a Neural Network (ALVINN) system. The model structure was relatively simple, comprising a fully-connected network, which is tiny by today’s standard. The network predicted actions from pixel inputs applied to simple driving scenarios with few obstacles. However, it demonstrated the potential of neural networks for end-to-end autonomous navigation. 

\begin{figure}[!htb]
	\includegraphics[width=6cm]{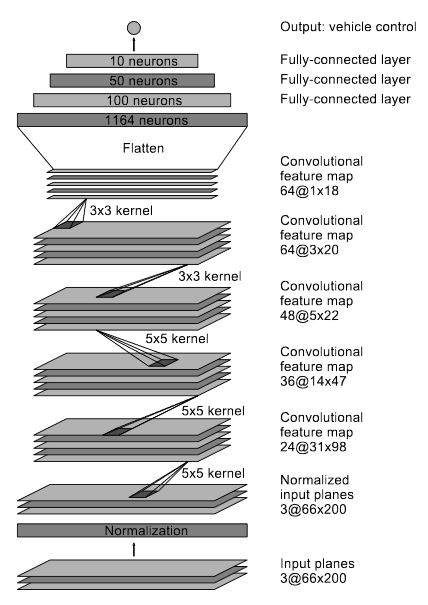}
	\centering
	\caption{CNN architecture used in \cite{bojarski2016end}. The network contains approximately 27 million connections and 250 thousand parameters.}
	\label{nvidiaimage}
\end{figure}

Last year, NVIDIA released a paper regarding a similar idea that benefited from ALVINN.  In the paper \cite{bojarski2016end}, the authors used a relatively basic CNN architecture to extract features from the driving frames. The layout of the architecture can be seen in Figure \ref{nvidiaimage}. Augmentation of the data collected was found to be important. The authors used artificial shifts and rotations of the training set. Left and right cameras with interpolated steering angles were also incorporated. This framework was successful in relatively simple real-world scenarios, such as highway lane-following and driving in flat, obstacle-free courses.  

Recently, more attempts on using deep CNNs and RNNs to tackle the challenges of video classification \cite{karpathy2014large}, scene parsing \cite{farabet2013learning}, and object detection \cite{szegedy2013deep} have stimulated the applications of more complicated CNN architectures in autonomous driving. "Learning Spatiotemporal Features with 3D Convolutional Networks" introduces how to construct 3D Convolutional Networks to capture spatiotemporal features in a sequence of images or videos \cite{Tran_2015_ICCV}. "Beyond Short Snippets: Deep Networks for Video Classification" describes two ways including using LSTM to deal with videos \cite{yue2015beyond}. "Deep Residual Learning for Image Recognition" \cite{he2016deep} and "Densely Connected Convolutional Networks" \cite{huang2016densely} describe the techniques to construct residual connections between different layers and make it easier to train deep neural networks.

Besides of the CNN and/or RNN methods, there are more research initiatives applying deep learning techniques in autonomous driving challenges. Another line of work is to treat autonomous navigation as a video prediction task. Comma.ai \cite{santana2016learning} has proposed to learn a driving simulator with an approach that combines a Variational Auto-encoder (VAE) \cite{kingma2013auto} and a Generative Adversarial Network (GAN) \cite{goodfellow2014generative}. Their approach is able to keep predicting realistic-looking video for several frames based on previous frames despite the transition model being optimized without a cost function in the pixel space. 

Moreover, deep reinforcement learning (RL) has also been applied to autonomous driving \cite{el2017deep}, \cite{shalev2016safe}. RL has not been successful for automotive applications until some recent work shows the deep learning algorithms’ ability to learn good representations of the environment. This was demonstrated by learning of games like Atari and Go by Google DeepMind \cite{mnih2013playing}, \cite{silver2016mastering}. Inspired by these work, \cite{el2017deep} has proposed a framework for autonomous driving using deep RL. Their framework is extensible to include RNN for information integration, which enables the car to handle partially observable scenarios. The framework also integrates attention models,  making use of the glimpse and action networks to direct the CNN kernels to the places of the input data that are relevant to the driving process.

\section{Methods}

We developed two types of models. The first one uses 3D convolutional layers followed by recurrent layers using LSTM. The second model uses transfer learning with 2D convolutional layers on a pre-trained model where the first layers are blocked from training.

\subsection{3D Convolutional Model with Residual Connections and Recurrent LSTM Layers}

\subsubsection{3D Convolutional Layer}
How 3D convolutional layer works is similar to 2D convolutional layers, the only difference is that in addition to height and width, now we have the third dimension depth (temporal). Instead of having a 2D filter (if we ignore the channel dimension for a while) moving within the image along height and width, now we have a 3D filter moving along with height, width and depth. If the input has shape ($D_1$, $H_1$,  $W_1$, $C$), then the output would have shape ($D_2$, $H_2$, $W_2$, $F$) where $F$ is the number of filters. $D_2$, $H_2$, $W_2$ could be calculated given stride and padding in its dimension.

\subsubsection{Residual Connection}
Since deep neural networks may have issues passing the gradient through all the layers, residual connections are used to help the training process. The idea of residual connection is to use network layers to fit a residual mapping instead of directly trying to fit a desired underlying mapping. 
Without residual connection:\\
Fit $H(G(x))$ directly\\
With residual connection:\\
Fit $F(x)=H(G(x))-G(x)-x$\\
Essentially the gradient can skip over different layers. This skipping lowers the effective number of layers the gradient has to pass through in order to make it all the way back. Skipping in this manner alleviates problems with the backpropagation algorithm in very deep networks such as the vanishing gradient problem.

\subsubsection{Spatial Batch Normalization}
Batch Normalization (see \cite{ioffe2015batch}) alleviates a lot of headaches with properly initializing neural networks by explicitly forcing the activations throughout a network to take on a unit gaussian distribution at the beginning of the training. Spatial batch normalization not only normalizes among different samples but also among the spatial axis of images. Here we add spatial batch normalization after each 3D convolutional layer.

\subsubsection{Recurrent Neural Networks and Long Short Term Memory}
Recurrent Neural Networks have loops in them, which allows information to persist. This could be used to understand present frame based on previous video frames. One problem of vanilla RNN is it cannot learn to connect the information too far away due to the gradient vanishing. LSTM has a special cell state, served as a conveyor belt that could allow information to flow without many interactions. After we use 3D convolutional layers to extract visual features, we feed them into LSTM layers to capture the sequential relation.

\subsubsection{New Architecture}
\begin{figure}[!htb]
	\includegraphics[width=4cm]{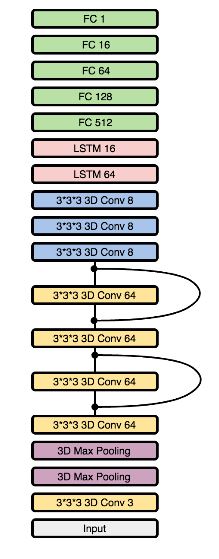}
	\centering
	\caption{3D convolutional model with residual connections and recurrent LSTM layers}
	\label{3dconvlstm_graph}
\end{figure}

For self driving cars, incorporating temporal information could play an important role in production systems. For example, if the camera sensor is fully saturated looking at the sun, knowing the information of the previous frames would allow for a much better prediction than basing the steering angle prediction only on the saturated frame. As discussed earlier, 3D convolutional layers and recurrent layers incorporate temporal information. In this model we combined these two ways of using temporal information. We used the idea residual connection in constructing this model \cite{he2016deep}. These connections allow for more of the gradient to pass through network by combining different layers. The model consisted five sequences of five frames of video shifted by one frame for the input (5x5x120x320x3). The values were selected to fit the computational budget. This allowed for both motion and differences in outputs between frames to be used. This model had 543,131 parameters. The architecture of the model can be seen in Figure \ref{3dconvlstm_graph}.

The model consists of a few initial layers to shrink the size followed by ResNet like blocks of 3D convolutions with spatial batch normalization (only two of these in the trained model). Due to computational restraints, shrink layers were added to make the input to the LSTM layers much smaller. Only two levels of recurrent layers were used due to the speed of computation on these layers being much slower due to parts that must be done in a serial manner. The output of the recurrent layers was fed into a fully connected stack that ends with the angle prediction. All of these layers used rectified linear units, ReLUs, as their activation except the LSTM layers (ReLUs keep the same positive value as the output and negative values are set to zero). Spatial batch normalization was used on the convolutional layers. The LSTM layers used the hyperbolic tangent function as their activation, which is common to use in these types of layers.

\subsection{Transfer Learning}
For this model, we used the idea of transfer learning. Transfer learning is a way of using high quality models that were trained on existing large datasets. The idea of transfer learning is that features learned in the lower layers of the model are likely transferable to another dataset. These lower level features would be useful in the new dataset such as edges.

\begin{figure}[!htb!]
	\includegraphics[width=4cm]{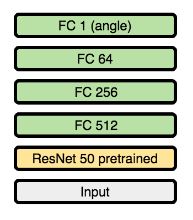}
	\centering
	\caption{Architecture used for transfer learning model.}
	\label{resnet50}
\end{figure}

Of the pre-trained models available, ResNet50 had good performance for this dataset. This model was trained on ImageNet. The weights of the first 15 ResNet blocks were blocked from updating (first 45 individual layers out of 175 total). The output of ResNet50 was connected to a stack of fully connected layers containing 512, 256, 64, and 1 different units respectively. The architecture of this model can be seen in Figure \ref{resnet50} with the overall number of parameters being 24,784,641. The fully connected layers used ReLUs as their activation. The ResNet50 model consists of several different repeating blocks that form residual connections. The number filters varies from 64 to 512. A block is consistent of a convolutional layer, batch normalization, ReLU activation repeated three times and the input layer output combined with the last layer.

Other sizes of locking were attempted, but produced either poor results or were slow in training. For example, training only the last 5 blocks provided poor results, which were only slightly better than predicting a steering angle of zero for all inputs. Training all the layers also produced worse results on the validation set compared to blocking the first 45 (0.0870 on the validation set after 32 epochs). 

The model took as input images of 224x224x3 (downsized and mildly stretched from the original Udacity data). The only augmentation provided for this model was mirrored images. Due to the size constraints of the input into ResNet50, cropping was not used as it involved stretching the image. The filters in the pretrained model were not trained on stretched images, so the filters may not activate as well on the stretched data (RMSE of 0.0891 on the validation set after 32 epochs). Additionally, using the left and the right cameras from the training set proved not to be useful for the 32 epochs used to train (0.17 RMSE on the validation set).

\section{Dataset and Features}
The dataset we used is provided by Udacity, which is generated by NVIDIAs DAVE-2 System \cite{bojarski2016end}. Specifically, three cameras are mounted behind the windshield of the data-acquisition car. Time-stamped video from the cameras is captured simultaneously with the steering angle applied by the human driver. This steering command is obtained by tapping into the vehicle's Controller Area Network (CAN) bus. In order to make the system independent of the car geometry, they represent the steering command as $1/r$ where $r$ is the turning radius in meters. They use $1/r$ instead of $r$ to prevent a singularity when driving straight (the turning radius for driving straight is infinity). $1/r$ smoothly transitions through zero from left turns (negative values) to right turns (positive values). Training data contains single images sampled from the video, paired with the corresponding steering command ($1/r$).

Training data set contains 101397 frames and corresponding labels including steering angle, torque and speed. We further split this data set into training and validation in a 80/20 fashion. And there is also a test set which contains 5615 frames. The original resolution of the image is 640x480.

Training images come from 5 different driving videos:\\

\begin{enumerate}
	\item 221 seconds, direct sunlight, many lighting changes. Good turns in beginning, discontinuous shoulder lines, ends in lane merge, divided highway
	\item discontinuous shoulder lines, ends in lane merge, divided highway
	791 seconds, two lane road, shadows are prevalent, traffic signal (green), very tight turns where center camera can't see much of the road, direct sunlight, fast elevation changes leading to steep gains/losses over summit. Turns into divided highway around 350s, quickly returns to 2 lanes.
	\item 99 seconds, divided highway segment of return trip over the summit
	\item 212 seconds, guardrail and two lane road, shadows in beginning may make training difficult, mostly normalizes towards the end
	\item 371 seconds, divided multi-lane highway with a fair amount of traffic
	
\end{enumerate}

\begin{figure}[!htb]
	\includegraphics[width=.1\textwidth]{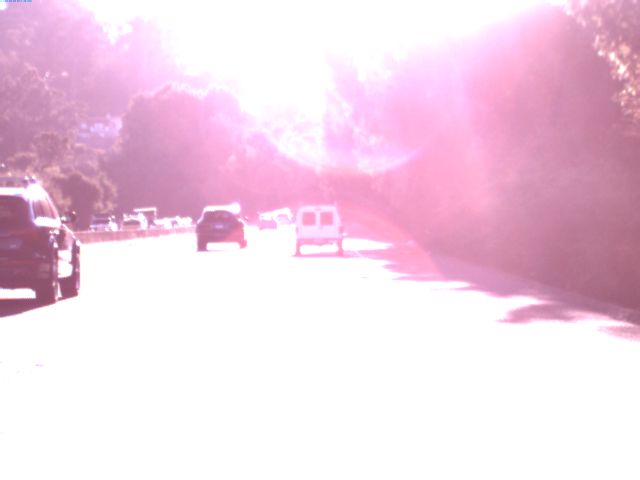}
	\includegraphics[width=.1\textwidth]{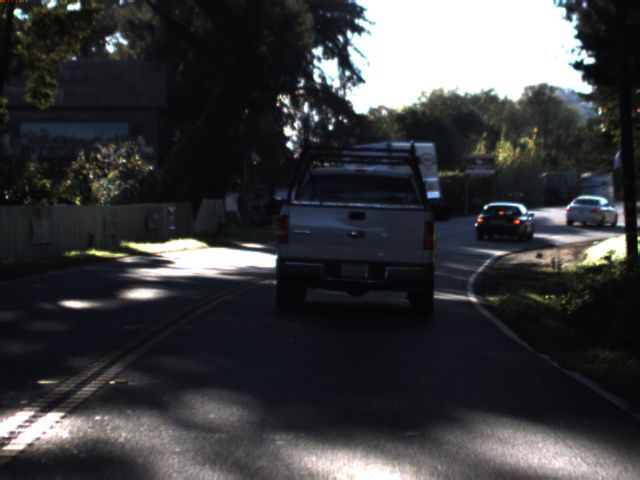}
	\includegraphics[width=.1\textwidth]{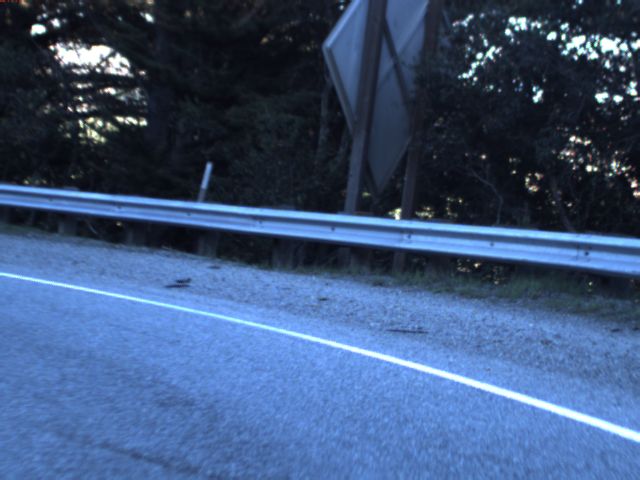}
	\includegraphics[width=.1\textwidth]{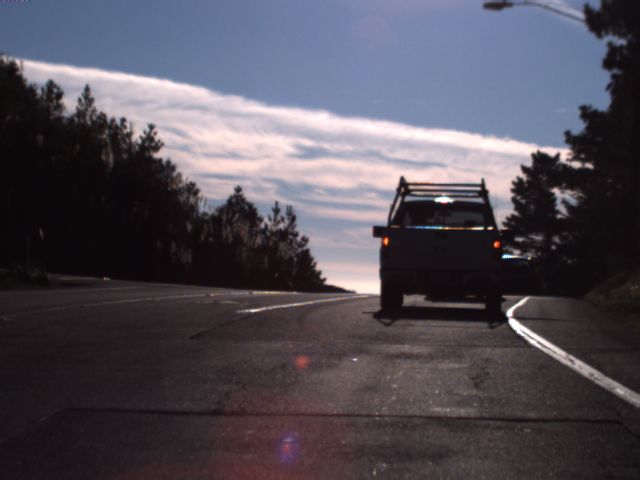}
	\includegraphics[width=.1\textwidth]{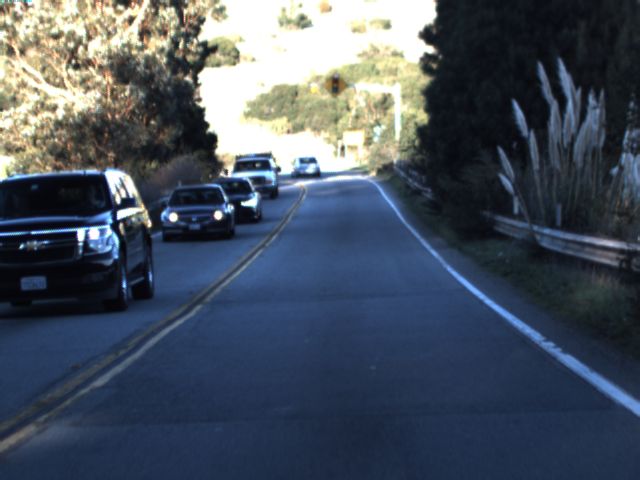}
	\includegraphics[width=.1\textwidth]{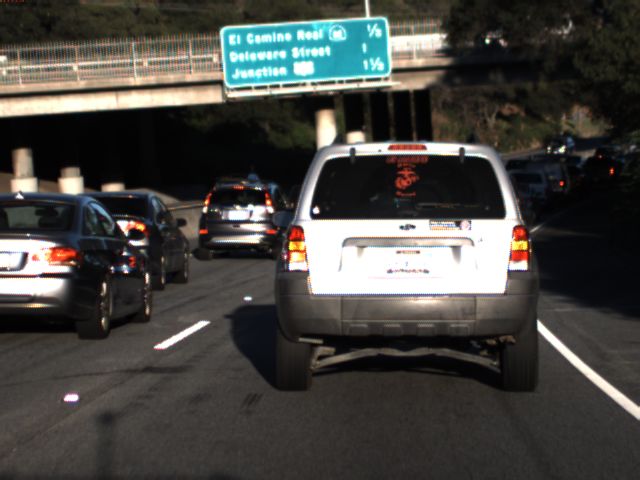}
	\centering
	\caption{Example images from the dataset. From left to right, bight sun, shadows, sharp left turn, up hill, straight, and heavy traffic conditions.}
	\label{example_images}
\end{figure}

Figure \ref{example_images} shows typical images for different light, traffic and driving conditions.

\subsection{Data Augmentation Methods}

\subsubsection{Brightness Augmentation}
Brightness is randomly changed to simulate different light conditions. We generate augmented images with different brightness by first converting images to HSV, scaling up or down the V channel and converting back to the RGB channel. Following are typical augmented images.

\begin{figure}[!htb]
	\includegraphics[width=8cm]{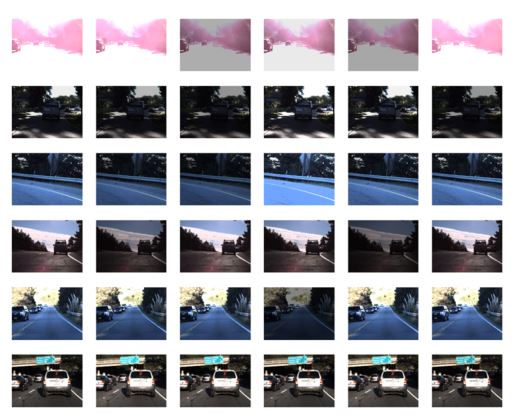}
	\centering
	\caption{Brightness augmentation examples}
	\label{bright_aug}
\end{figure}

\subsubsection{Shadow Augmentation}
Random shadows cast across images. The intuition is that even the camera has been shadowed (maybe by rainfall or dust), the model is still expected to predict the correct steering angle. This is implemented by choosing random points and shading all points on one side of the image.

\begin{figure}[!htb]
	\includegraphics[width=8cm]{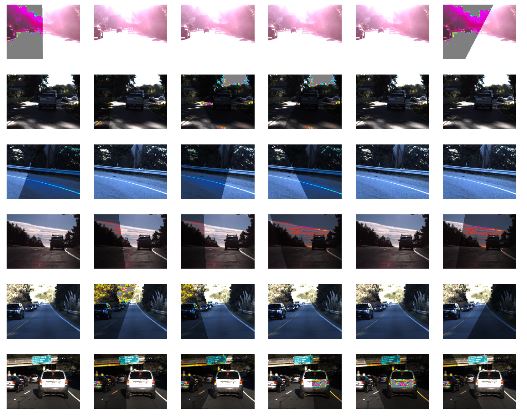}
	\centering
	\caption{Shadow augmentation examples}
	\label{shadow_aug}
\end{figure}

\subsubsection{Horizontal and Vertical Shifts}
We shift the camera images horizontally to simulate the effect of car being at different positions on the road, and add an offset corresponding to the shift to the steering angle. We also shift the images vertically randomly to simulate the effect of driving up or down a slope.

\begin{figure}[!htb]
	\includegraphics[width=8cm]{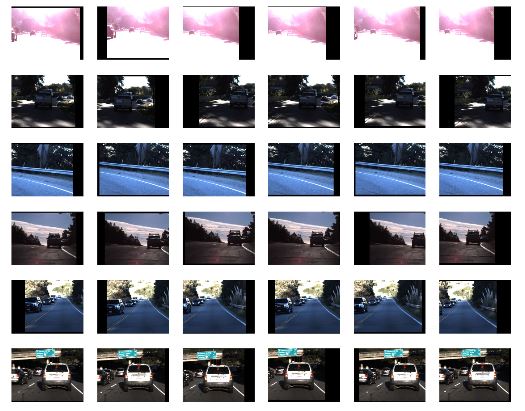}
	\centering
	\caption{Shift augmentation examples}
	\label{shift_aug}
\end{figure}

\subsubsection{Rotation Augmentation}
During the data augmentation experiment, the images were also rotated around the center. The idea behind rotations is that the model should be agnostic to camera orientation and that rotations may help reduce over-fitting.

\begin{figure}[!htb]
	\includegraphics[width=4cm]{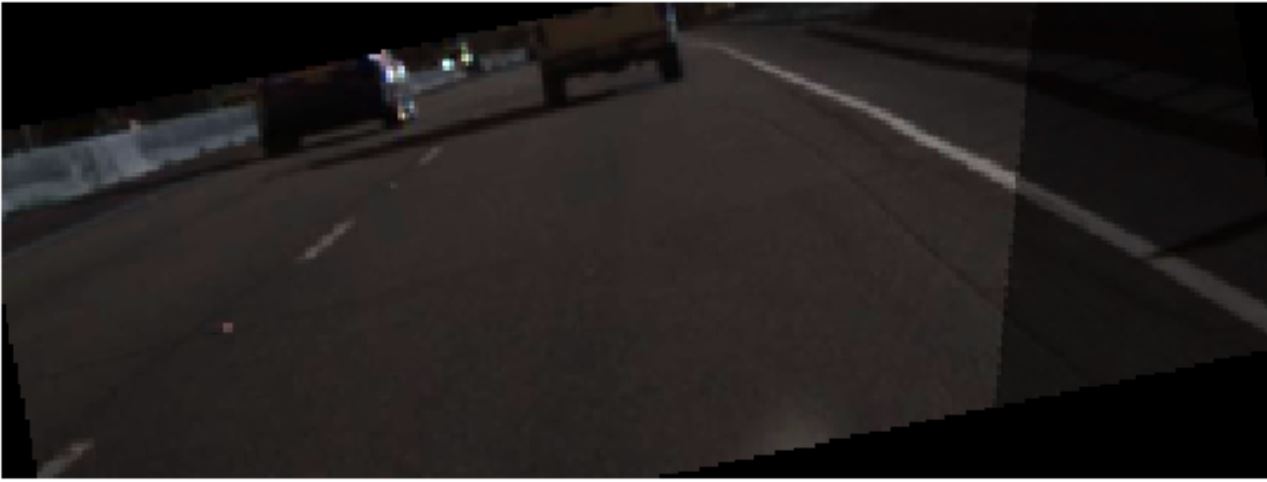}
	\centering
	\caption{Rotation augmentation example}
	\label{rotaion_aug}
\end{figure}

\subsection{Preprocessing}
For each image, we normalize the value range from [0, 255] to [-1, 1] by normalizing by image=-1+2*original images/255. We further rescale the image to a 224x224x3 square image for transfer learning model. For the 3D LSTM model, we cropped the sky out of the image to produce data the size of 120x320x3.

\section{Experiments, Results, and Discussion}

\subsection{Data Augmentation}
In order to establish a baseline for our research, we used the architecture from NVIDIA \cite{bojarski2016end} to test different forms of data augmentation. Teams in the Udacity challenge noted that data augmentation was helpful along with the original NVIDIA researchers. Knowing which forms of augmentation work well for the amount of time and computational available would be helpful in training our new models. The NVIDIA model architecture seen previously in Figure \ref{nvidiaimage}. The input to this model was 120x320x3 with a batch size of 32. In the NVIDIA paper \cite{bojarski2016end} it was not clear how they optimized the loss function. For this experiment, we used the default parameters of Adam (see \cite{kingma2014adam}) provided in Keras (learning rate of 0.001, $\beta_{1}=0.9$, $\beta_{2}=0.999$, $\epsilon=1e-8$, and decay=0).

Three different levels of augmentation were examined. The first had minimal augmentation with only using random flips and cropping of the top of the image. Randomly flipping the input images eliminates the bias towards right turns found in the dataset. Cropping the top of the image eliminates the sky from the image, which should not play a role in how to turn predict the steering angle. A second form of augmentation had the same augmentation as the minimal version along with small rotations (-5 to +5 degrees), shifts (25 pixels), and small brightness changes of the image. The final Heavier version of augmentation used more exaggerated effects of the second version including large angle rotations (up to 30 degrees), large shadows, shifts, and larger brightness changes were used. Results from this experiment can be see in Table \ref{data_aug_table}.

\begin{table}[!htb]
	\centering
	\caption{RMSE on the validation set using the NVIDIA architecture for different levels of data augmentation with 32 epochs.}
	\label{data_aug_table}
	\begin{tabular}{|l|l|l|}
		\hline
		\textbf{Minimal} & \textbf{Moderate} & \textbf{Heavy} \\ \hline
		0.09             & 0.10              & 0.19           \\ \hline
	\end{tabular}
\end{table}
 
Using heavy data augmentation produced very poor results that were not much above predicting a steering angle of 0 for all the data. The moderate augmentation produced good results; however the minimal augmentation produced the best results. These results could be explained by only training for 32 epochs. Heavy augmentation could be hard for the model to pick up on such drastic shifts. Similarly, the moderate version may have outperformed the minimal version over more epochs. In a later section visualization of these tests will be examined. For our new models, we chose to use minimal augmentation.

\subsection{Training Process}

\subsubsection{Loss Function and Optimization}
For all models used, the mean-square-loss function was used. This loss function is common for regression problems. The MSE punishes large deviations harshly. This function is simple the mean of the sum of the squared differences between the actual and predicted results (see Equation \ref{mse}). The scores for the Udacity challenge were reported as the root-mean-square-error, RMSE, which is simply the square root of the MSE.

\begin{equation} \label{mse}
\begin{split}
MSE=\frac{1}{n} \sum (y_{i}-\hat{y_{i}})^{2}\\
RMSE=\sqrt{\frac{1}{n} \sum (y_{i}-\hat{y_{i}})^{2}} 
\end{split}
\end{equation}

To optimize this loss, the Adam optimizer was used \cite{kingma2014adam}. This optimizers is often the go to choice for deep learning application. This optimization method usually substantially outperforms more generic stochastic gradient decent methods. Initial testing of these models indicate that their loss level change slowed after a few epochs. Although Adam computes an adaptive learning rate through its formula, decay of the learning rate was used. The decay rate of the optimizer was updated from 0 to the learning rate divided by the number of epochs. The other default values of the Keras Adam optimizer showed good results during training (learning rate of 0.001, $\beta_{1}=0.9$, $\beta_{2}=0.999$, $\epsilon=1e-8$, and decay=learning rate/batch size).

\subsection{Feature Visualization}

In order to examine what our networks find relevant in an image, we can use saliency maps. These maps can show how the gradient flows back to the image highlighting the most salient areas. A similar approach was used in a recent NVIDIA paper \cite{bojarski2017explaining}.

\subsubsection{Data Augmentation Experiment}
\begin{figure}[!htb]
	\includegraphics[width=4cm]{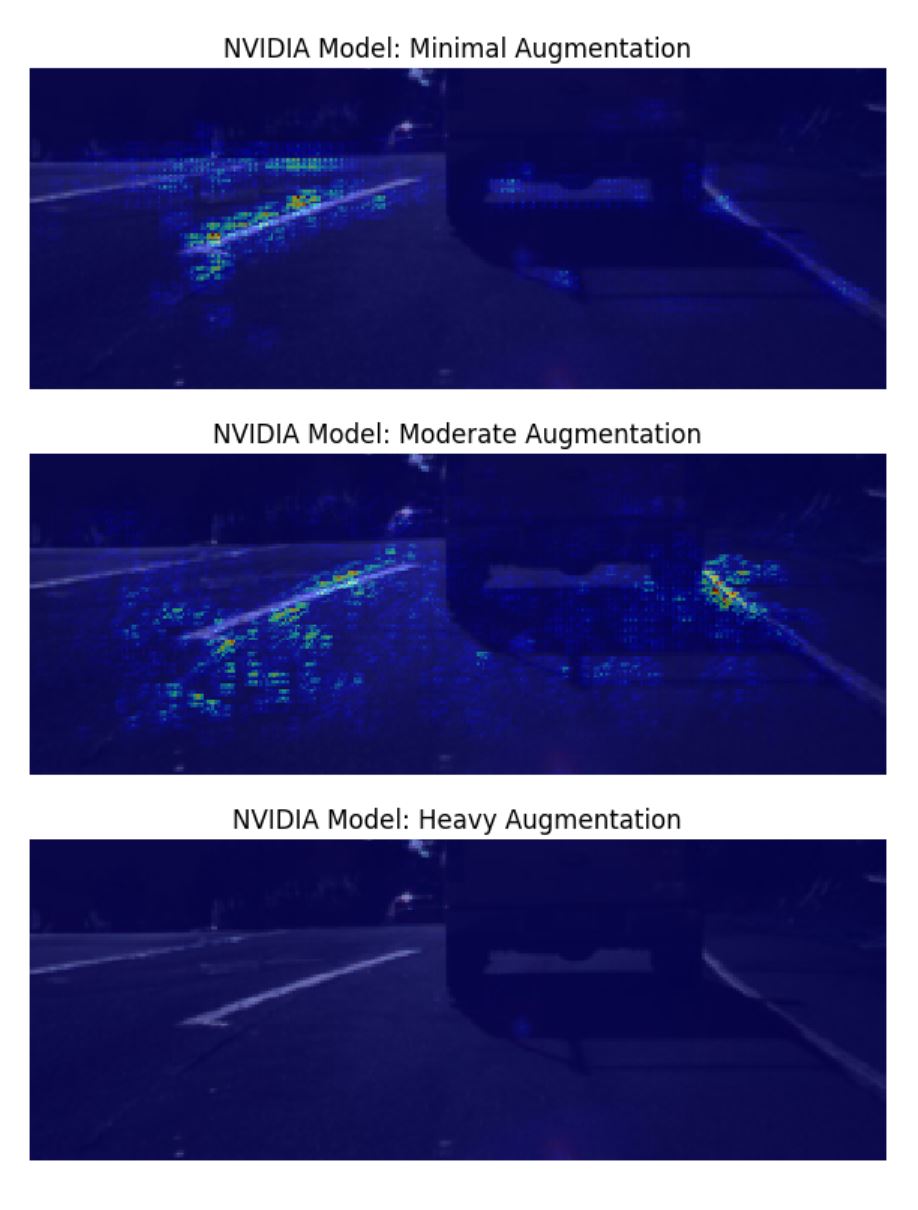}
	\centering
	\caption{NVIDIA model saliency maps for different levels of augmentation.}
	\label{nvidia_aug}
\end{figure}

What these models found important can be visualized in Figure \ref{nvidia_aug}. The minimal model found the lane markers important. In the moderate model more of the lane markers were found to be important; however, this model's saliency maps appeared more noisy, which could explain its slightly decreased performance. In the heavy model almost no areas were found to be salient, which is understandable due to its poor performance. 

\subsubsection{3D Convolutional LSTM Model}

\begin{figure}[!htb]
    \includegraphics[width=4cm]{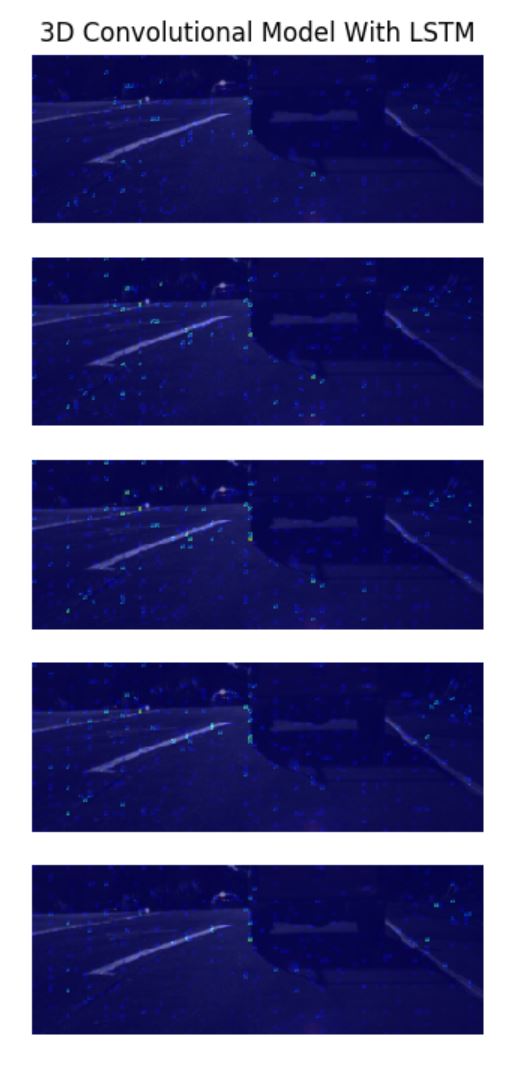}
    \centering
    \caption{Saliency map for a sequence in the 3D convolutional LSTM model for an example image.}
    \label{3d_full}
\end{figure}

This model produced interesting saliency maps. In examining on of the video clips fed into the model, we can see that the salient features change frame to frame in Figure \ref{3d_full}. The salient features seem to change from frame to frame, which would indicate that the changes between frames are important.

\begin{figure}[!htb]
	\includegraphics[width=8cm]{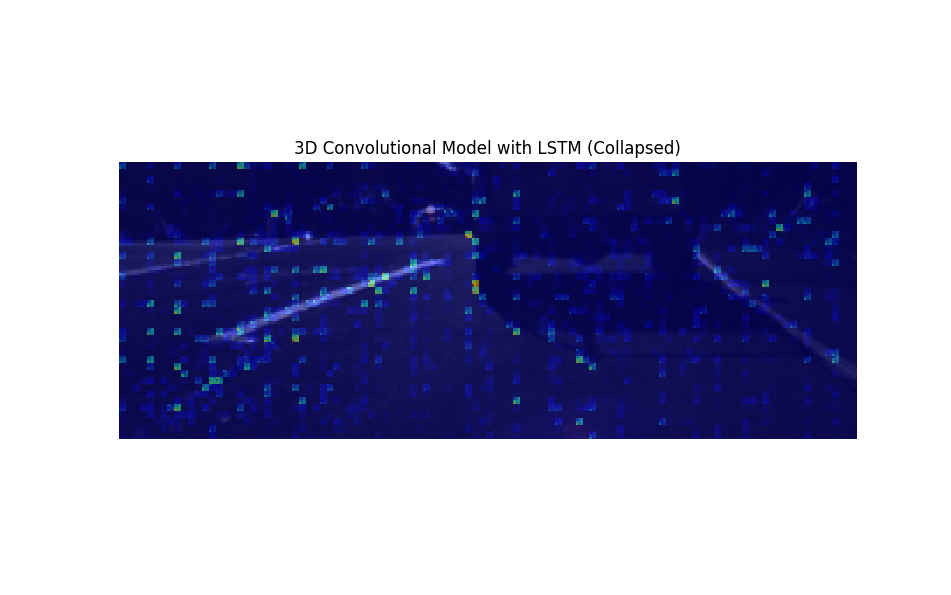}
	\centering
	\caption{Saliency map for the 3D convolutional LSTM model for an example image.}
	\label{3d_collapse}
\end{figure}

 This sequence of frames can be collapse into a single image, which is shown in Figure \ref{3d_collapse}. The collapsed version helps to visualize this better. The expressed salient features do cluster around road markers, but they also cluster around other vehicles and their shadows. This model may be using information about the car in front in order to make a steering angle prediction along with the road markers. 

\subsubsection{Transfer Learning Model}

\begin{figure}[!htb]
	\includegraphics[width=8cm]{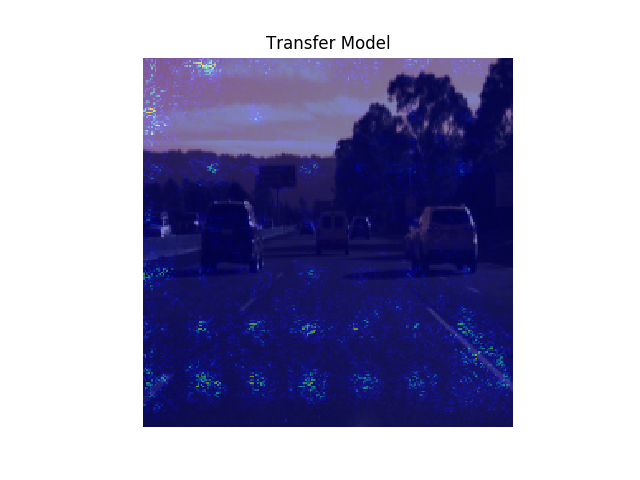}
	\centering
	\caption{Transfer learning model (ResNet50) saliency map for an example image.}
	\label{resnet50_saliency}
\end{figure}

An example saliency map for the ResNet50 transfer learning model can be seen in Figure \ref{resnet50_saliency}. The model does appear to have salient features on the road markers; however, there are also regularly spaced blotches. These blotches may be artifacts from using this type pretrained model with residual connections. Although this model had the best overall results, its saliency maps did not match well with the expectation of what would be expected for salient features in predicting steering angles from road images. 

\subsection{Results}
These models were all ran on the same datasets (training, validation, and test). The results for each model is listed in Table \ref{model_results}. The results from the 3D convolutional model with LSTM and residual connections had a RMSE for the test set of 0.1123. Since the Udacity challenge is over, the results can be compared to the leader board. For the 3D LSTM model, the results on the test set would have put in in tenth place overall. The ResNet50 transfer model had the best results overall with a RMSE of 0.0709 on the test set. This result would have placed the model in fourth place overall in the challenge. This is without using any external functions for the models (some teams used an external smoothing function in conjunction with their deep learning models).

\begin{table}[!htb]
	\centering
	\caption{RMSE for the models on the Udacity dataset.}
	\label{model_results}
	\begin{tabular}{|l|l|l|l|}
		\hline
		& \textbf{Training Set} & \textbf{Validation Set} & \textbf{Test Set} \\ \hline
		\textbf{Predict 0}      & 0.2716                & 0.2130                  & 0.2076            \\ \hline
		\textbf{3D LSTM}        & 0.0539                & 0.1139                  & 0.1123            \\ \hline
		\textbf{Transfer}       & 0.0212                & 0.0775                  & 0.0709            \\ \hline
		\textbf{NVIDIA}         & 0.0750                & 0.0995                  & 0.0986            \\ \hline
	\end{tabular}
\end{table}

\begin{figure}[!htb]
	\includegraphics[width=8cm]{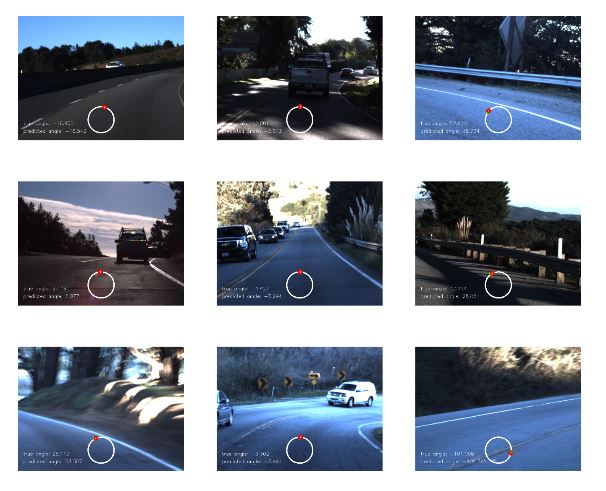}
	\centering
	\caption{Example actual vs. predicted angle on unprocessed images (transfer model).}
	\label{angle_overlay}
\end{figure}

In order to help visualization of what this level of error looks like, an example overlay for random images is seen in Figure \ref{angle_overlay}. The green circle indicates the true angle and the red circle indicates the predicted angle. The predictions are generated from the ResNet50 transfer learning model.

\subsection{Discussion}
For the amount of epochs we used, only minimal data augmentation proved to be of any major use for these model. For more expansive training, the strategy of data augmentation can allow for near infinite training data given the right strategy. 

Overall, these models showed that they were competitive with other top models from the Udacity challenge. For the 3D LSTM model, with more time and computational resources, this model could have been expanded to take in a longer period of video along with more ResNet blocks. Expanding the model in this way could have produced superior results. One of the teams near the top of the competition used a full 250 frames or 2.5 seconds of video \cite{udacity}.

For the Resnet50 transfer model, the strategy of using a pre-trained model, locking approximately the first quarter layers, training the deeper layers with the existing weights, and connecting to a fully connected stack proved to be effective in producing a high quality and competitive model for the Udacity self-driving car dataset. It was surprising that this model outperform the other model. The architecture of this model takes no temporal data, yet it still predicts very good values.

Both of these models appeared to have had some over-fitting with the ResNet50 model having more of an issue with this. Data augmentation could act as a form of regularization for this model. Different teams in the Udacity challenge have tried different regularization method including dropout and $L_{2}$ regularization. The results for using this regularization methods was mixed with some teams claiming good results and others having less success.

\section{Conclusion and Future Work}

In examining the final leader board from Udacity our models would have placed fourth (transfer learning model) and tenth (3D convolutional model with LSTM layers). These results were produced solely from the models without any external smoothing function. We have shown that both transfer learning and a more advanced architecture have promise in the field of autonomous vehicles. The 3D model was limited by computational resources, but overall it still provided a good result from a novel architecture. In future work the 3D model's architecture could be expanded by having a larger and deeper layers, which may produce better results.

These models are far from perfect and there is substantial research that still needs to be done on the subject before models like these can be deployed widely to transport the public. These models may benefit from a wider range of training data. For a production system, a model would have to be able to handle the environment in snowy conditions. Generate adversarial models, GANs, could be used to transform a summer training set into a winter one. Additionally, GANs could be used to generate more scenes with sharp angles. Additionally, a high quality simulator could be used with deep reinforcement learning. A potential reward function could be getting from one point to another while minimizing time, maximizing smoothness of the ride, staying in the correct lane/following the rules of the road, and not hitting objects.

{\small
\bibliographystyle{ieee}
\bibliography{egbib}
}

\end{document}